\documentclass{article}
\usepackage{spconf,amsmath,graphicx,multirow}

\usepackage{makecell}

\title{Efficient Uncertainty Estimation with Gaussian Process \\for Reliable Dialog Response Retrieval}
\name{Tong Ye$^{1,2}$,  Zhitao Li$^{1}$, Jianzong Wang$^{1,*}$\thanks{$^*$Corresponding author: Jianzong Wang, jzwang@188.com.}, Ning Cheng$^{1}$, Jing Xiao$^{1}$}
\address{$^{1}$Ping An Technology (Shenzhen) Co., Ltd.\\
	$^{2}$University of Science and Technology of China}
\begin{document}
%\ninept
%
\maketitle
\begin{abstract}
Deep neural networks have achieved remarkable performance in retrieval-based dialogue systems, but they are shown to be ill calibrated. Though basic calibration methods like Monte Carlo Dropout and Ensemble can calibrate well, these methods are time-consuming in the training or inference stages. To tackle these challenges, we propose an efficient uncertainty calibration framework GPF-BERT for BERT-based conversational search, which employs a Gaussian Process layer and the focal loss on top of the BERT architecture to achieve a high-quality neural ranker. Extensive experiments are conducted to verify the effectiveness of our method. In comparison with basic calibration methods, GPF-BERT achieves the lowest empirical calibration error (ECE) in three in-domain datasets and the distributional shift tasks, while yielding the highest $R_{10}@1$ and MAP performance on most cases. In terms of time consumption, our GPF-BERT has an 8$\times$ speedup.
\end{abstract}
\section{Introduction}
\label{sec:intro}

Dialog response retrieval models based on deep neural networks (DNNs) primarily focus on modeling the relevance between context and responses and have achieved impressive performance \cite{devlin2018bert,penha2021calibration,cohen2021not}.  However, these models always suffered from over or under confidence due to the poor calibration of DNNs \cite{guo2017calibration}. As a result, it is difficult to determine whether the predictions are reliable. 
This attribute is essential for distribution shifts tasks and safety-critical areas since erroneous predictions can result in far more significant consequences than not making any prediction at all \cite{DBLP:conf/acl/KamathJL20}.
Therefore, an ideal dialog model should exhibit confidence in its predictions while also recognizing situations where its predictions may be incorrect and uncertain.

Uncertainty modeling has been touched in previous work on dialog systems. 
Monte Carlo (MC) Dropout \cite{penha2021calibration,cohen2021not,gal2016dropout} and Ensemble \cite{penha2021calibration,durasov2021masksembles} have emerged as two of the most prominent uncertainty estimation methods for deep retrieval networks. While Ensemble trains independently multiple models using stochastic gradient descent, MC Dropout trains a single stochastic network by dropping different subsets of weights simultaneously in train and test time \cite{durasov2021masksembles}. Unfortunately, MC Dropout necessitates carrying out several forward passes and Ensemble becomes computationally expensive. This poses a significant challenge, particularly in light of the widespread adoption of large transformer architectures like BERT \cite{durasov2021masksembles}. Therefore, it is urgent to explore an efficient method to quantify the uncertainty in deep neural retrieval models.

Gaussian Process (GP) \cite{williams2006gaussian} is a well-established framework for evaluating uncertainty. As an input moves farther away from the training data, the level of uncertainty in GP predictions tends to increase \cite{dutordoir2021deep}. 
However, GP is challenging to scale to large datasets and improve the performance while DNNs are computationally scalable enough to handle them \cite{dutordoir2021deep}.
SNGP \cite{liu2020simple} combines the strengths of GP and DNNs, utilizing spectral normalization \cite{miyato2018spectral} to the weights in each residual layer, which can efficiently handle the large scale inputs and makes robust uncertainty-aware predictions. Unfortunately, the application of SNGP has not been explored in the retrieval-based dialog system.

So motivated, we attempt to investigate a simple and efficient approach for the well-calibrated dialog response retrieval models based on Gaussian Process. Specifically, we add a neural GP layer to a deterministic BERT-like backbone to improve the ability of uncertainty estimation and train the model with focal loss \cite{lin2017focal} to achieve better calibration. Different from MC Dropout and Ensemble, our method only needs to be performed by passing through a single forward so that GPF-BERT achieves almost 8$\times$ speedup in terms of inference time. 
To summarize, the main contributions of this work are as follows:

\begin{itemize}
	
	\item To our best knowledge, we first estimate uncertainty in dialog tasks with SNGP. Furthermore, we propose an efficient framework GPF-BERT to estimate uncertainty combining the focal loss and SNGP. 
	\item We conduct extensive experiments to compare the performance of various calibration methods. Our method yields the lowest ECE in three in-domain datasets and the distributional shift task while keeping performance.

\end{itemize}

	\begin{figure}
		\centering
		\includegraphics[width=0.9\linewidth,height=0.12\textheight]{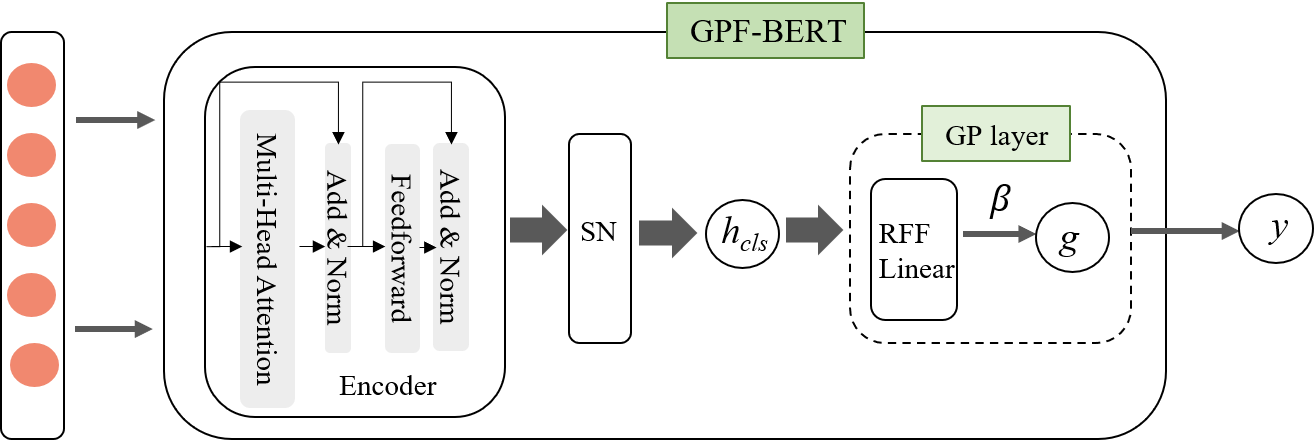}
		\caption{An illustration of GPF-BERT prediction models for dialog response retrieval.}
		\vspace{-0.2cm}
		\label{fig1}
	\end{figure}

\section{METHODOLOGY}
In this section, we present details of GPF-BERT, utilizing a neural Gaussian Process layer to model uncertainty as shown in Fig \ref{fig1}.

\subsection{Model Architecture}
For a dialog dataset, we denote a training set as a triples $\{(U_i, r_i, y_i)\}_{i=1}^N$, where $U_i=\{u_{i1}, u_{i2}, \ldots, u_{it}\}$ is a dialog context consisting of $t$ utterances and $y_i$ is the response relevance label $y_i\in \{0,1\}$, with a response candidate $r_i$. The whole dialog context with the candidate is fed into BERT-like encoders, $x_i=$$\{[CLS]$, $u_{i1}$, $\ldots$, $u_{it}$, [SEP]$, r_i\}$, where the special token $[CLS]$ denotes the sequence beginning and $[SEP]$ separates the response from the contexts.
For each training example, the representation vector $h(x_i)$ of $[CLS]$ is the feature of this dialog.

For a latent representation $h_i=h(x_i)$, the GP output layer $g_i =g(h_i)$ follows a multivariate normal distribution a prior: $g \sim GP(0,K)$, where $K$ is a $N\times N$ kernel matrix. For more details of GPs, refer to \cite{williams2006gaussian}. 

First, spectral normalization (SN) \cite{miyato2018spectral} involves decomposing the parameters $W$ of each neural network layer using SVD and subsequently constraining the maximum singular value to 1. Briefly, the SN estimates the spectral norm $\left\| W_l \right\| _2$ using the power iteration method during each training step and subsequently normalizes the weights according to the estimated norm as follows: 
\begin{equation}
	w_l=\begin{cases}
		c*W_l/\left\| W_l \right\| _2 & \quad if \ c<\left\| W_l \right\| _2\\
		W_l & \quad     otherwise\\
	\end{cases}
\end{equation}

To approximate the kernel matrix with a low-rank approximation, GPF-BERT uses a technique known as random Fourier features (RFF) \cite{rahimi2007random} as  $K=\Phi \Phi^T(\Phi \in R^{N\times L})$, where $L$ denotes the dimensionality of the latent space:
\begin{equation}
	\begin{aligned}
		&g \sim GP(0,\Phi \Phi^T),
		\Phi_i =\sqrt{2/L}*cos(-Wh_i+b)\label{2}
	\end{aligned}
\end{equation}
$\Phi_i$ contains a fixed weight matrix $W$ with entries sampled from $N(0,1)$, and a fixed bias vector $b$ with entries sampled from $U(0,2\pi)$.

Finally, the GP layer as a neural network layer \cite{lee2017deep} with learnable output weights $\beta \sim N(0,I)$ according to the RFF approximation to the GP prior.
\begin{equation}
	g(h_i)=\sqrt{2/L}*cos(-Wh_i+b)^T\beta\label{3}
\end{equation}	

Note that the use of random Fourier features (RFF) to linearly transform the GP helps to overcome the curse of dimensionality and yield good performance with finite data. However, this technique may also result in feature collapse, which can compromise model robustness, particularly when dealing with distributional shifts. Therefore, we deploy SN on the weights, which enforces the feature extractor to be bi-Lipschitz to mitigate feature collapse \cite{van2021feature}.

Due to the lack of conjugacy between the classified likelihood function and a Gaussian prior, we resort to using the Laplace approximation to quantify uncertainty \cite{liu2020simple}. The maximum a posteriori (MAP) solution is denoted by $\hat{\beta}$ and the Laplace posterior for GP under the RFF approximation can be expressed as:
\begin{equation}
	\begin{aligned}
		&p(\beta|D)\sim GP(\hat{\beta},
		\hat{\Sigma}_k
		),
		\begin{matrix}
			\hat{\Sigma}_{k}^{-1} \end{matrix}=\begin{matrix} \sum_{i=1}^{N} \hat{p_i}(1-\hat{p_i})\Phi_i \Phi_i^T \end{matrix} +I\label{4}
	\end{aligned}
\end{equation}
During minibatch training, $\hat{\beta}$ is updated via regular SGD with respect to the loss function and $\hat{\Sigma}_{t}^{-1}$ is updated cheaply using
\begin{equation}
	\begin{aligned}
		\hat{\Sigma}_{t}^{-1}  =\alpha* 
		\hat{\Sigma}_{t-1}^{-1}  + (1-\alpha)*\begin{matrix}\sum_{i=1}^{M} \hat{p_i}(1-\hat{p_i})\Phi_i \Phi_i^T \end{matrix} \label{5}
	\end{aligned}
\end{equation}
where $t$ indexes update steps, $M$ is the mini-batch size, $\hat{p_i}$ is the softmax probability and $\alpha$ is a small scaling coefficient.

For a given feature vector $x^*$ of a query-response pair, GPF-BERT computes the posterior mean $\hat{m}(x^*)=\Phi^T\beta$ and the variance $\hat{K}(x^*)=\Phi^T\hat{\Sigma}\Phi$. Finally, the predictive distribution is written as $p=exp(m)/\sum_i exp(m_i)$ where $m \sim N(\hat{m},\hat{K})$ and we calculate its posterior mean using mean-field approximation \cite{lu2020mean} for lower computation.

\subsection{Loss Function}
The conventional cross-entropy loss assigns the same weight to individual samples on a mini-batch. However, there are several low-confidence samples hard to classify. Although the high-confidence samples have a small loss, their cumulative loss value is still greater than the low-confidence samples due to the large number, dominating gradient and producing bad performance. That is just an important reason for poor calibration of cross-entropy \cite{mukhoti2020calibrating}. 

In this work, our GPF-BERT utilizes focal loss \cite{lin2017focal}, which focuses more on the uncertain samples by reducing the weight of high-confidence samples.
Namely, by reducing the weight of the easy samples, the model focuses more on the hard samples when training. 
\begin{equation}
	\begin{aligned}	
		L_{focal}&=-(1-\hat{p})^\gamma log(\hat{p})
		\ge -(1-\gamma \hat{p})log\hat{p}\\
		&=L_{ce}-\gamma H[\hat{p}]
		\ge KL(q||\hat{p})-\gamma H[\hat{p}]
	\end{aligned}
\end{equation}

When training, focal loss ensures minimization of the KL divergence whilst simultaneously increasing the entropy $H[\hat{p}]$ \cite{mukhoti2020calibrating}. The high entropy can help prevent the model from becoming overconfident and thereby improve calibration.
We use $L_{ce}$ and $L_{focal}$ to respectively denote the cross-entropy loss and the focal loss with hyperparameter $\gamma \ge 1$.

\section{experiments}
\subsection{Experiments Setup}
\noindent
\textbf{Datasets:} We utilize three large-scale conversational response ranking datasets in our experiments.
MS Dialog \cite{qu2018analyzing} contains 246,000 context-response pairs culled from over the Microsoft Answer community.  
MANtIS \cite{penha2019introducing} contains 1.3 million context-response pairs including 14 different domain.	
Ubuntu Dialogue Corpus v1.0 (UDC) \cite{lowe2015ubuntu} is consisted of almost 1 million context-response pairs.

\noindent
\textbf{Metrics:} We use $R_{10}@1$ and MAP to measure the retrieval performance and ECE \cite{naeini2015obtaining} for calibration. 
We divide the interval $[0,1]$ into $M=10$ equispaced bins. 
The ECE calculate a weighted average of the absolute difference between the accuracy $A_i$ and confidence $B_i$ of each bin: $ECE=\sum_{i=1}^{M} \frac{|B_i|}{N}|mean(A_i)-mean(B_i)|$.

\noindent
\textbf{Baselines:} We compare BERT \cite{devlin2018bert} with other common calibrated methods.
MC Dropout \cite{penha2021calibration} approximates Bayesian inference using dropout during training and testing, and produces a predictive distribution by performing multiple forward passes. In this paper, we use 10 forward passes based on a BERT model.
Ensemble \cite{penha2021calibration} independently trained several models and integrate their predictions  a model integrating predictions of BERT and MC Dropout models. 
SNGP \cite{liu2020simple} adds a weighting normalization step during training and replaces the dense output layer with a GP layer.

\noindent
\textbf{Implementation Details:}
We use 12-layered $BERT$ as the backbone, each encoder having 12 attention heads and a hidden dimension of 768. All the methods are obtained a latent representation by extracting the $[CLS]$ feature. The dropout probability was set to 0.1 and the learning rate was set to 5e-6 using the Adam optimizer. Following recent research \cite{penha2021calibration} that employed finetuned BERT for dialog response ranking, we randomly select nine responses from the list of all responses as the negative samples when training. The hyperparameters utilized for the deterministic variant are also employed for each model architecture. In our experiment, we first train our model for 1 epochs with a batch size of 16 on a cluster of 1 Tesla V100 with 16G memory. Simultaneously, all the experiment results are the average over 5 runs along with the standard error. 

\begin{table}[]
	
	\caption{Calibration (ECE) and effectiveness ($R_{10}@1$, MAP). $"\uparrow"$ represents higher is better and $"\downarrow"$ means lower is better.}
	\scalebox{0.85}{
		\begin{tabular}{ll|ccc}
			\Xhline{1.2pt}
			&  & $R_{10}@1$$\uparrow$ & MAP$\uparrow$ & ECE$\downarrow$ \\ \Xhline{1.2pt}
			\multirow{5}{*}{\rotatebox{90}{MSDialog}} & BERT  &\textbf{0.682$\pm$0.006}     &0.800$\pm$0.003     &0.125$\pm$0.020     \\
			& MC Dropout   &0.673$\pm$0.005     &0.796$\pm$0.003     &0.110$\pm$0.020     \\
			& Ensemble    &0.680$\pm$0.004     &\textbf{0.800$\pm$0.003}     &0.115$\pm$0.019     \\ \cline{2-5}
			& SNGP  &0.659$\pm$0.013     &0.783$\pm$0.008     &0.110$\pm$0.006     \\
			& GPF-BERT  &0.681$\pm$0.006     &0.799$\pm$0.003     &\textbf{0.025$\pm$0.010}     \\ \Xhline{1.2pt}
			\multirow{5}{*}{\rotatebox{90}{MANtIS}}   & BERT  &0.590$\pm$0.012     &0.713$\pm$0.010     &0.169$\pm$0.028     \\ 
			& MC Dropout   &0.591$\pm$0.011     &0.713$\pm$0.009     &0.152$\pm$0.02     \\
			& Ensemble    &0.592$\pm$0.011     &0.713$\pm$0.010     &0.157$\pm$0.026     \\ \cline{2-5}
			& SNGP  &0.597$\pm$0.021     &0.719$\pm$0.013     &0.147$\pm$0.010     \\
			& GPF-BERT  &\textbf{0.614$\pm$0.017}    &\textbf{0.729$\pm$0.012}     &\textbf{0.025$\pm$0.009}     \\ \Xhline{1.2pt}
			\multirow{5}{*}{\rotatebox{90}{UDC}}      & BERT  &0.810$\pm$0.003     &0.880$\pm$0.002     &0.037$\pm$0.001     \\ 
			& MC Dropout   &0.809$\pm$0.003     &0.878$\pm$0.002     &0.033$\pm$0.001     \\
			& Ensemble    &0.810$\pm$0.003     &0.879$\pm$0.002     &0.034$\pm$0.001   \\ \cline{2-5}
			& SNGP  &0.806$\pm$0.001     &0.877$\pm$0.001     &0.033$\pm$0.001    \\ 
			& GPF-BERT  &\textbf{0.818$\pm$0.001}     &\textbf{0.885$\pm$0.001}     &\textbf{0.016$\pm$0.002}     \\ \Xhline{1.2pt}
	\end{tabular}}
	\vspace{-0.3cm}	
	\label{tab:my-table1}
\end{table} 

\begin{table*}[]
	\setlength{\tabcolsep}{3mm}{
		\caption{Calibration (ECE) and effectiveness ($R_{10}@1$, MAP) for distributional shift tasks. "$\uparrow$" represents higher is better and "$\downarrow$" means lower is better. All the models are trained in one dataset and test in the other dataset.}
		\label{tab:my-table2}
		
		\scalebox{0.92}{
			\begin{tabular}{cllccccc}
				\Xhline{1.2pt}
				\multicolumn{1}{l}{Train}  & Test                       & Metric & BERT & MC Dropout & Ensemble & SNGP & GPF-BERT \\ \Xhline{1.2pt}
				\multirow{6}{*}{MS Dialog}  
				& \multirow{3}{*}{MANtIS}                     & $R_{10}@1$$\uparrow$     &0.378$\pm$0.024         &0.357$\pm$0.017         &0.369$\pm$0.020$\pm$          &\textbf{0.385$\pm$0.040}      &0.381$\pm$0.033     \\
				&                            & MAP$\uparrow$     &0.538$\pm$0.018            &0.524$\pm$0.012         &0.533$\pm$0.015          &\textbf{0.543$\pm$0.030}      &0.540$\pm$0.023     \\
				&                            & ECE$\downarrow$     &0.343$\pm$0.035             &0.328$\pm$0.045         &0.331$\pm$0.037          &0.307$\pm$0.011      &\textbf{0.206$\pm$0.054}     \\ \cline{2-8}
				& \multirow{3}{*}{UDC}       & $R_{10}@1$$\uparrow$     &0.609$\pm$0.009            &0.600$\pm$0.007         &0.606$\pm$0.008         &0.602$\pm$0.007      &\textbf{0.612$\pm$0.014}     \\
				&                            & MAP$\uparrow$     &0.736$\pm$0.005             &0.730$\pm$0.004         &0.734$\pm$0.004          &0.731$\pm$0.004      &\textbf{0.737$\pm$0.009}     \\
				&                            & ECE$\downarrow$     &0.109$\pm$0.011             &0.092$\pm$0.011         &0.097$\pm$0.009          &0.085$\pm$0.003      &\textbf{0.026$\pm$0.016}     \\\Xhline{1.2pt}
				\multirow{6}{*}{MANtIS}    & \multirow{3}{*}{MS Dialog} & $R_{10}@1$$\uparrow$     &0.430$\pm$0.069             &0.418$\pm$0.060         &0.427$\pm$0.065          &0.363$\pm$0.108      &\textbf{0.452$\pm$0.037}     \\
				&                            & MAP$\uparrow$     &0.598$\pm$0.058             &0.591$\pm$0.050         &0.596$\pm$0.054          &0.548$\pm$0.089      &\textbf{0.619$\pm$0.026}     \\
				&                            & ECE$\downarrow$     &0.514$\pm$0.037             &0.497$\pm$0.040         &0.503$\pm$0.037          &0.485$\pm$0.017      &\textbf{0.364$\pm$0.019}     \\ \cline{2-8}
				& \multirow{3}{*}{UDC}                        & $R_{10}@1$$\uparrow$     &0.662$\pm$0.008            &0.660$\pm$0.010         &\textbf{0.662$\pm$0.009}          &0.656$\pm$0.013      &0.659$\pm$0.015     \\
				&                            & MAP$\uparrow$     &0.769$\pm$0.008          &0.768$\pm$0.006         &\textbf{0.769$\pm$0.006}          &0.766$\pm$0.009      &0.768$\pm$0.009     \\
				&                            & ECE$\downarrow$     &0.071$\pm$0.006             &0.061$\pm$0.006         &0.064$\pm$0.006          &0.059$\pm$0.008      &\textbf{0.025$\pm$0.011}     \\ \Xhline{1.2pt}
				\multirow{6}{*}{UDC}       & \multirow{3}{*}{MS Dialog} & $R_{10}@1$$\uparrow$     &0.324$\pm$0.031             &0.290$\pm$0.065         &0.312$\pm$0.072          &0.406$\pm$0.092      &\textbf{0.431$\pm$0.074}     \\
				&                            & MAP$\uparrow$     &0.513$\pm$0.072            &0.484$\pm$0.064         &0.503$\pm$0.069          &0.581$\pm$0.072      &\textbf{0.603$\pm$0.059}     \\
				&                            & ECE$\downarrow$     &0.616$\pm$0.031             &0.622$\pm$0.026         &0.610$\pm$0.030          &0.607$\pm$0.034      &\textbf{0.528$\pm$0.042}     \\ \cline{2-8}
				& \multirow{3}{*}{MANtIS}                     & $R_{10}@1$$\uparrow$     &0.250$\pm$0.045             &0.220$\pm$0.038         &0.240$\pm$0.044          &0.270$\pm$0.027      &\textbf{0.283$\pm$0.035}     \\
				&                            & MAP$\uparrow$     &0.418$\pm$0.037             &0.389$\pm$0.032         &0.408$\pm$0.036          &0.437$\pm$0.021      &\textbf{0.447$\pm$0.029}     \\
				&                            & ECE$\downarrow$     &0.510$\pm$0.041            &0.537$\pm$0.037         &0.508$\pm$0.040          &0.494$\pm$0.032      &\textbf{0.400$\pm$0.045}     \\ \Xhline{1.2pt}
	\end{tabular}}}
	\vspace{-0.2cm}

\end{table*}	
\begin{table}[]
	\vspace{-0.4cm}
	\setlength{\tabcolsep}{0.8mm}{
		\caption{Parameters and inference time.}
		\label{tab:my-table3}
		\scalebox{0.85}{
			\begin{tabular}{lcccc}
				\Xhline{1.2pt}
				models     & BERT   & MC Dropout & Ensemble & GPF-BERT \\ \Xhline{1.2pt}
				Parameters(M) & 413.26 & 413.26     & 826.52   & 453.55   \\
				Time(min)       & 6.80(1$\times$)    & 65.00(9.56$\times$)         & 71.80(10.56$\times$)     & 7.87(1.16$\times$)   \\ \Xhline{1.2pt}
	\end{tabular}}}	
	\vspace{-0.3cm}
\end{table}

\subsection{Results}
\noindent
\textbf{In-domain.} Table \ref{tab:my-table1} reports the results of GPF-BERT and all baselines on the MS dialog,  MANtIS and UDC datasets. From this table, GPF-BERT generally outperforms other single-model approaches in ECE across various datasets with a reduction of less than 1\% in $R_{10}@1$ and MAP.

Specifically, BERT, which is a vanilla model without any calibration, usually performs well but is not well calibrated. The calibration of MC Dropout and Ensemble exceeds BERT, which verifies that Bayesian models exhibit greater expressiveness in their ability to convey confidence, but unfortunately still obtain poor calibration. Compared to the prior methods, the ECE of GPF-BERT is almost lowest, which is reduced by almost 10\%, 14\% and 2\% respectively in three datasets, while the $R_{10}@1$ and MAP are better or less than 1\% decrease. Namely, GPF-BERT includes uncertainty information while keeping $R_{10}@1$ and MAP performance in in-domain datasets. 

\noindent
\textbf{Distributional Shift.}
In addition to using the uncertainty estimation for in-domain datasets, we also train the model using the training set from one dataset, i.e. train set, and evaluate it on a different dataset's test set, which is also known as domain generalization or distributional shift tasks. We record all the retrieval performance and calibration in Table \ref{tab:my-table2}. As shown, we observe that SNGP was reduced by up to 4\% when compared to the BERT, MC Dropout and Ensemble.
Moreover, the GPF-BERT achieves a substantial decrease up to 10\% to the upper calibration bound under the SNGP framework, even though the $R_{10}@1$ and MAP of GPF-BERT is lower in distribution shift tasks. 
This confirms that GP-based retrieval models will have the more robust expressiveness to convey confidence in distributional shift tasks. According to the results of SNGP and GPF-BERT, we find that the focal loss plays an important role in calibration.

\noindent
\textbf{Efficiency.}
One of the most critical challenges to overcome is the computational cost when employing Bayesian to capture uncertainty. We analyze the efficiency of GPF-BERT in terms of parameter number and inference time in the MS Dialog dataset in Table \ref{tab:my-table3}. Compared to MC Dropout and Ensemble, GPF-BERT has a significant decrease (at least 8 times) in inference time. While not completely free, GPF-BERT only adds negligible computational cost, but it greatly improves the calibration, which facilitates adaptation to other models. 

We believe that GP maintains a distribution over functions rather than model parameters, which enables GPF-BERT to improve uncertainty calibration for dialog response retrieval models. In addition, according to a recent theorem  \cite{kristiadi2020being} that capturing uncertainty information and correcting overconfidence can be achieved by making only the last layer of a model in binary classification, we can assume that adding a GP layer is Bayesian enough so that GPF-BERT can achieve better calibration.

\subsection{Ablation Study}
\begin{figure}
	\centering
	\vspace{-0.1cm}
	\includegraphics[width=0.9\linewidth,height=0.19\textheight]{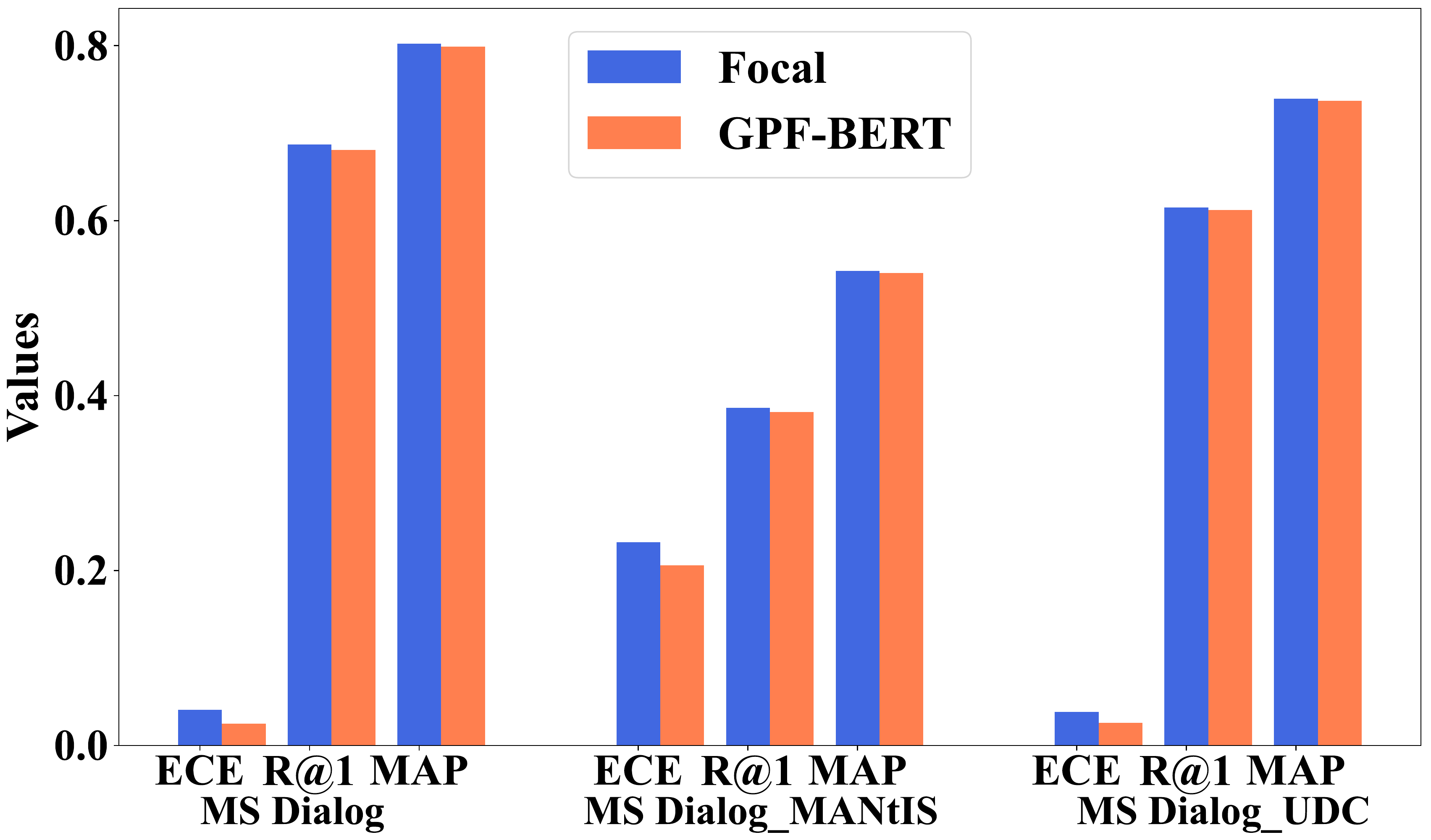}
	\vspace{-0.4cm}		
	\caption{Ablation study.}
	\vspace{-0.5cm}
	\label{ablation}
\end{figure}

To understand the impact of focal loss on calibration improvement, we conducted a straightforward ablation study on the in-domain dataset MS Dialog and its two distribution shift tasks as shown in Fig \ref{ablation}. Obviously, the $R_{10}@1$ and MAP of GPF-BERT are lower than Focal but the calibration is much better. Focal loss is a commonly used regularization method and we found that it still facilitates calibration when applied to SNGP. On the other hand, the difference in architecture between Focal and GPF-BERT demonstrates that the GP-based model architecture is a better framework for calibration than conventional models.
That is to say, the model architecture may be key to improving calibration. 

\section{conclusion}
In this paper, we present an efficient uncertainty estimation architecture GPF-BERT for reliable dialog response retrieval tasks. GPF-BERT only adds a neural GP layer to a deterministic DNN to improve the ability of uncertainty estimation and trains the model with focal loss to achieve better calibration while maintaining the flexibility of deep neural networks. We conducted extensive experiments to verify the effectiveness including parameters and inference time. Furthermore, we explored the relative contributions of focal loss to the effectiveness improvement in the ablation study.

\section{Acknowledgement}
This paper is supported by the Key Research and Development Program of Guangdong Province under grant No. 2021B0101400003. 
Jianzong Wang from Ping An Technology (Shenzhen) Co., Ltd (jzwang@188.com) is the corresponding author.

\bibliographystyle{IEEEbib}
\bibliography{refs}

\end{document}